\documentclass{article}

\usepackage{arxiv}

\usepackage[utf8]{inputenc}
\usepackage[T1]{fontenc}
\usepackage{hyperref}
\usepackage{url}
\usepackage{booktabs}
\usepackage{amsfonts}
\usepackage{nicefrac}
\usepackage{microtype}
\usepackage{lipsum}
\usepackage{graphicx}
\usepackage[numbers, sort&compress]{natbib}
\usepackage{mathtools}
\usepackage{algorithm}
\usepackage{algpseudocode}
\usepackage{multicol}
\usepackage{multirow}
\usepackage[para]{threeparttable}
\usepackage{wrapfig}
\usepackage{amsmath}
\usepackage{appendix}

\newtheorem{theorem}{Theorem}

\title{Harnessing the Power of Local Representations for Few-Shot Classification}

\date{}

\author{
  Shi Tang \\
  School of Software \\
  Tsinghua University \\
  \And
  Guiming Luo\thanks{Corresponding author, \href{mailto:gluo@tsinghua.edu.cn}{gluo@tsinghua.edu.cn}.\\
  \hspace*{2em}This work was supported by the National Natural Science Foundation of China (NSFC) No.62173203.
  } \\
  School of Software \\
  Tsinghua University \\
  \AND
  Xinchen Ye \\
  DUT-RU ISE \\
  Dalian University of Technology \\
  \And
  Zhiyi Xia \\
  School of Software \\
  Tsinghua University \\
}

\renewcommand{\headeright}{Preprint}
\renewcommand{\undertitle}{}

\begin{document}
\maketitle

\begin{abstract}
  Generalizing to novel classes unseen during training is a key challenge of few-shot classification.
  Recent metric-based methods try to address this by local representations. However, they are unable to take full advantage of them due to (i) improper supervision for pretraining the feature extractor, and (ii) lack of adaptability in the metric for handling various possible compositions of local feature sets.
  In this work, we harness the power of local representations in improving novel-class generalization.
  For the feature extractor, we design a novel pretraining paradigm that learns randomly cropped patches by soft labels. It utilizes the class-level diversity of patches while diminishing the impact of their semantic misalignments to hard labels.
  To align network output with soft labels, we also propose a UniCon KL-Divergence that emphasizes the equal contribution of each base class in describing ``non-base'' patches.
  For the metric, we formulate measuring local feature sets as an entropy-regularized optimal transport problem to introduce the ability to handle sets consisting of homogeneous elements.
  Furthermore, we design a Modulate Module to endow the metric with the necessary adaptability.
  Our method achieves new state-of-the-art performance on three popular benchmarks.
  Moreover, it exceeds state-of-the-art transductive and cross-modal methods in the fine-grained scenario.
\end{abstract}

\keywords{Few-shot classification \and Metric learning \and Meta-learning}

\section{Introduction}

Given abundant samples of some classes (often called base classes) for training, few-shot classification (FSC) aims at distinguishing between novel classes unseen during training with limited examples.
Suffering from the low-data regimes and the inconsistency between training with base classes and inference on novel classes, FSC algorithms often struggle with poor generalization to novel classes.

To address this, recent approaches \cite{EMD,FRN,random_walk,EMD2} resort to local representations.
Specifically, an image is represented by a set of local features instead of a global embedding, in the hope of providing transferrable information across categories through possible common local features.
Then, a set metric is employed to measure image relevance for classification following metric learning \cite{NIPS2002_c3e4035a}.
Evidently, the quality of the feature extractor and the set metric are crucial.
However, both aspects are unsatisfactory in existing methods, resulting in an unexploited potential of local representations in improving novel-class generalization.

\begin{figure}[tb]
    \centering
    \includegraphics[width=\linewidth]{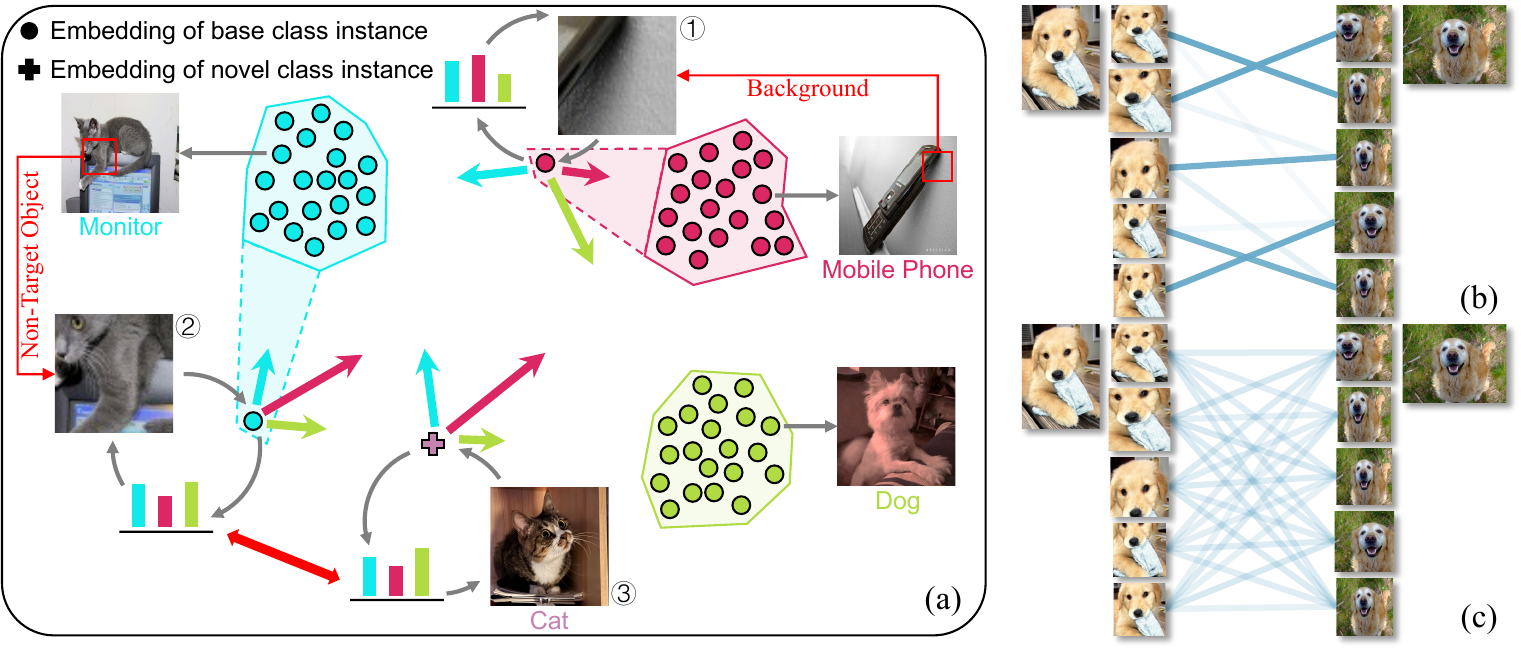}
    \caption{(a) Hard labels could provide false supervision since random cropping may alter the semantics. Describing patches by analogy, soft labels can avoid this and utilize the class-level diversity provided by random cropping. The matching flows between two sets of similar local patches using (b) EMD and (c) our Adaptive Metric.}
    \label{fig:beginning}
\end{figure}

\textbf{Feature extractor is insufficiently pre-trained with hard labels only.}
Usually, the encoder for extracting local features will be pre-trained with a proxy task classifying all base classes, where random cropping is often inherited as a simple and effective augmentation \cite{meta_baseline,random_walk,EMD2,PMF}.
However, it may alter the semantics (Fig.~\ref{fig:beginning}~(a)~\raisebox{.5pt}{\textcircled{\raisebox{-.9pt} {1}}}\raisebox{.5pt}{\textcircled{\raisebox{-.9pt} {2}}}), making ground-truth hard labels insufficient for pretraining encoders capable of extracting high-quality local features.
The reasons are, firstly, they may provide false supervision that correlates the background or non-target objects to a base class.
This is acceptable for normal intra-class classification tasks as it serves as a shortcut knowledge (e.g., dolphins are usually in the water) which improves the performance \cite{xiao2021noise}. But in the few-shot setting, these priors do not hold for novel classes, which introduces bias.
Secondly, they cannot utilize the class-level diversity provided by random cropping to prevent the network from overfitting to base classes. Because hard labels strictly assume that the input belongs to one of the base classes, which cannot describe patches with semantics beyond all base classes.

Indicating the probability of the input belonging to each base class, soft labels are capable of describing cropped patches by analogy\footnote{The reason for this is that patch features can be represented linearly or nonlinearly by the manifold base \cite{robust_PCA} which is instantiated as mean features of base classes here.} (e.g., a cat is something more like a dog and less like a monitor).
Therefore, they can be used to supervise the learning of these cropped patches for regularization while avoiding false supervision.
Moreover, soft labels connect non-target objects with potential novel classes through similar distributions (Fig.~\ref{fig:beginning}~(a)~\raisebox{.5pt}{\textcircled{\raisebox{-.9pt} {2}}}\raisebox{.5pt}{\textcircled{\raisebox{-.9pt} {3}}}), making the learning of these patches a pre-search for suitable positions to embed novel class samples, which warms up the encoder for possible test scenarios in advance.

\textbf{Metric necessitates adaptability for various set compositions.}
Given two sets to be measured, a local relation exists for each element in their Cartesian product. The rational utilization of these local relations is essential for an effective measure.
Recently, embedding randomly cropped patches stands out in constructing local feature sets \cite{EMD2} since it handles the uncertainty of novel classes with randomness.
However, random cropping unavoidably results in various set compositions, i.e., the local features in a set may be highly similar or different to varying degrees.
Different compositions need different local relation utilization, requiring the metric to be able to handle them adaptively.
For example, similar features of the same set yield similar local relations which should be utilized equivalently, while dissimilar features yield disparate local relations which should be utilized differently.
As a classic representative of optimal transport (OT) distances, Earth Mover's Distance (EMD) is a well-studied set metric and applies to measuring local feature sets \cite{EMD}.
However, it lacks the adaptability for various set compositions.
Specifically, the optimum transport matrix is usually solved on a vertex of the transport polytope, resulting in a sparse matching flow regardless of the set composition.
When the set elements are highly similar, only a few local relations among several equally important ones contribute to the matching process (Fig.~\ref{fig:beginning}~(b)), affecting the fidelity of the similarity measure.

Introducing an entropic regularization, the Sinkhorn Distance \cite{sinkhorn} encourages smoother transport matrices, allowing ``one-to-many'' matching (Fig.~\ref{fig:beginning}~(c)) that better utilizes similar local relations.
Therefore, formulating the metric as an entropy-regularized OT problem can endow the algorithm with the ability to handle similar local relations.
Moreover, adaptability can be introduced into the metric by conditioning the regularization strength on set compositions.

In this paper, we propose a novel method, namely FCAM, for few-shot classification to harness the power of local representations in improving novel-class generalization.
To obtain better local features, we propose Feature Calibration to pre-train few-shot encoders.
It supervises the learning of cropped patches with soft labels produced by a momentum-updated teacher, utilizing the class-level diversity of them while avoiding false supervision.
In addition, by decomposing the classical KL-Divergence commonly used for soft label supervision, we find its inherent weighting scheme unsuitable for learning few-shot encoders as it implicitly assumes that the input must belong to a certain base class.
Therefore, we propose UniCon KL-Divergence (UKD) with a more suitable weighting scheme for the soft label supervision.
To measure local feature sets with various compositions, we propose Adaptive Metric that formulates the set measure problem as a regularized OT problem, where a Modulate Module is designed to adjust the regularization strength adaptively.
The proposed method achieves new state-of-the-art performance on three popular benchmarks. Moreover, it exceeds state-of-the-art transductive and cross-modal methods in the fine-grained scenario.
Our contributions are as follows:
\begin{itemize}
    \item We propose a novel pretraining paradigm for few-shot encoders that uses soft labels to utilize the class-level diversity provided by random cropping while avoiding improper supervision.
    \item We propose a UniCon KL-Divergence for the soft label supervision to correct an assumption of conventional KL-Divergence that does not hold true for the few-shot setting.
    \item We propose a novel metric capable of handling various compositions of local feature sets adaptively for local-representation-based FSC.
\end{itemize}

\section{Related Work}

\textbf{Metric-based few-shot classification.}
The literature exhibits significant diversity in the area of FSC \cite{matchingnet,protonet,relationnet,random_walk,EMD2,MAML,delta_encoder,lgm_net}, among which metric-based methods \cite{matchingnet,protonet,relationnet,random_walk,EMD2} are very elegant and promising.
The main idea is to meta-learn a representation expected to be generalizable across categories with a predefined \cite{matchingnet,protonet,random_walk,EMD,EMD2} or also meta-learned \cite{relationnet} metric as the classifier.
For example, Snell \textit{et al.} \cite{protonet} average the embeddings of congener support samples as the class prototype and leverages the Euclidean distance for classification. Sung \textit{et al.} \cite{relationnet} replace the metric with a learnable module to introduce nonlinearity.
To avoid congener image-level embeddings from being pushed far apart by the significant intra-class variations, recent approaches resort to local representations.
Generally, an instance is represented by a local feature set whose elements can be implemented as local feature vectors \cite{DN4,EMD,random_walk,EMD2} or embeddings of patches cropped grid-like \cite{random_walk,EMD2} or randomly \cite{EMD2}.
Then, support-query pairs are measured by a metric capable of measuring two sets, e.g., accumulated cosine similarities between nearest neighbors \cite{DN4}, bidirectional random walk \cite{random_walk} or EMD \cite{EMD,EMD2}.

\textbf{Self-distillation.}
First proposed for model compression \cite{model_compression,mimic_model,distill_hinton}, knowledge distillation aims at transferring ``knowledge'', such as logits \cite{distill_hinton} or intermediate features \cite{Yim_2017_CVPR,NEURIPS2018_6d9cb7de,Heo_2019_ICCV}, from a high-capability teacher model to a lightweight student network. As a special case when the teacher and student architectures are identical, self-distillation has been consistently observed to achieve higher accuracy \cite{pmlr-v80-furlanello18a}. Zhang \textit{et al.} \cite{self_distill_label_smoothing} relate self-distillation with label smoothing, a commonly-used regularization technique to prevent models from being over-confident.
Generating soft labels with a momentum-updated teacher, our Feature Calibration is closely related to self-distillation and exhibits a similar regularization effect for improving class-level generalization.

\textbf{Optimal transport distances.}
The distances based on the well-studied OT problem are very powerful for probability measures. EMD was first proposed for image retrieval \cite{1997The} and exhibited excellent performance. Cuturi \cite{sinkhorn} proposes the Sinkhorn Distance by regularizing the OT problem with an entropic term, which greatly improves the computing efficiency and defines a distance with a natural prior on the transport matrix: everything should be homogeneous in the absence of a cost.
It provides an essential prior that EMD lacks for measuring sets consisting of highly similar nodes.

\begin{figure}[tb]
    \centering
    \includegraphics[width=\linewidth]{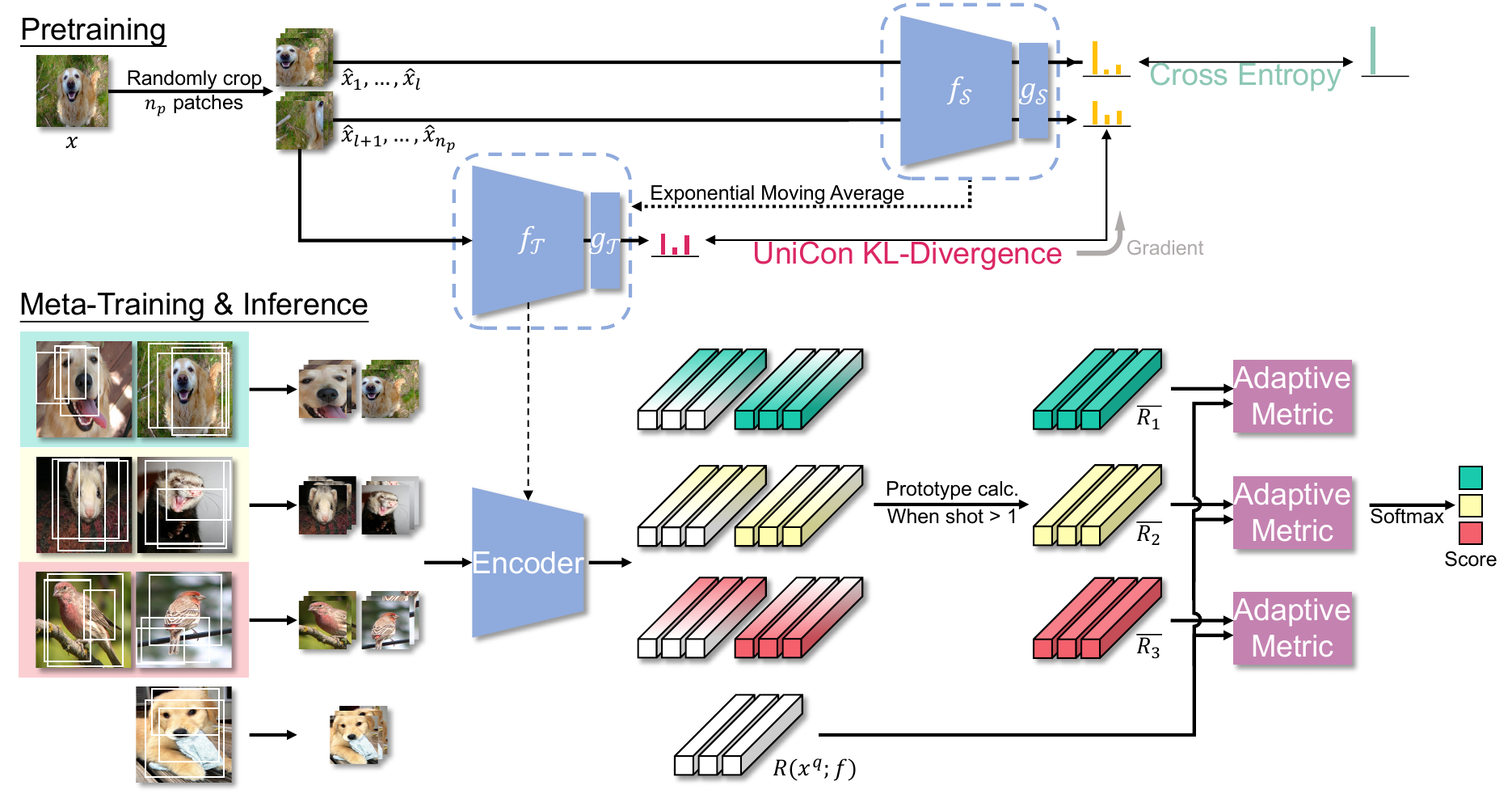}
    \caption{Overview of our framework ($3$-way $2$-shot as an example).}
    \label{fig:framework}
\end{figure}

\section{Training Paradigm}

Given a labeled dataset $D_{base}=\{(x_i^b, y_i^b)\}_{i=1}^{N_{base}}$ composed of $n_c$ base classes, few-shot classification aims to construct a classifier for future tasks consisting of novel classes.
Contemporary metric-based methods \cite{meta_baseline,random_walk,EMD2,PMF} often adopt a ``pretraining + meta-training'' paradigm.
In the pretraining stage, a classification network $\phi=f\circ g$ consisting of an encoder $f$ and a linear layer $g$ is trained to distinguish all base classes, i.e., $\phi(x_i^b)\in\mathbb{R}^{n_c}$.
In the meta-training stage, the encoder is fine-tuned across a large number of $N$-way $K$-shot tasks constructed from $D_{base}$ to simulate the test scenario.
In a task containing $N$ classes ($N<n_c$), $K$ samples from each class are sampled to construct a support set $D_{spt}=\{(x_i^s,y_i^s)\}_{i=1}^{NK}$, according to which we need to predict labels for a query set $D_{qry}=\{(x_i^q,y_i^q)\}_{i=1}^{NQ}$ that contains samples from the same $N$ classes with $Q$ samples per class.
Specifically, for a query sample $x_i^q$ whose ground-truth label $y_i^q=c$ ($c\in\{1,\ldots,N\}$), the pre-trained encoder $f$ is fine-tuned to maximize:
\begin{equation}\label{eq:metric}
    p(y_i^q=c\mid x_i^q)=\frac{\exp{(S(R(x_i^q;f),\overline{R_c}))}}{\sum\nolimits_{j=1}^N\exp{(S(R(x_i^q;f),\overline{R_j}))}},
\end{equation}
where $S(\cdot,\cdot)$ is a metric measuring the similarity between $x_i^q$'s representation $R(x_i^q;f)$ and class $j$'s prototype representation $\overline{R_j}$.

As illustrated in Fig.~\ref{fig:framework}, we integrate our method into this paradigm and focus on local representations constructed by the random cropping operation $\xi(\cdot)$, where $R(x_i^q;f)$ and $\overline{R_j}$\footnote{For cases where shot $K>1$, we conduct additional prototype calculation for a structured FC layer \cite{EMD2}.} are defined as:
\begin{equation}\label{eq:rep}
\begin{split}
    R(x_i^q;f)\coloneqq\{\mathbf{u}_m|m=1,2,\ldots,n\},\quad\overline{R_j}\coloneqq\{\mathbf{v}_m|m=1,2,\ldots,n\}, \\
    \text{where}~\mathbf{u}_m=f(\xi(x_i^q)),\quad\mathbf{v}_m=\frac{1}{K}\sum\nolimits_{i=1}^{NK}f(\xi(x_i^s))\cdot[y_i^s=j],
\end{split}
\end{equation}
where $[y_i^s=j]$ is an indicator function that equals $1$ when $y_i^s=j$ and $0$ otherwise.

\section{Feature Calibration}

\subsection{Feature Calibration with Soft Labels}

Different from existing methods using only ground-truth hard labels, we propose a novel paradigm that takes soft labels into account as well for pretraining few-shot encoders.
It calibrates the extracted features by avoiding false supervision while fully exploiting the class-level diversity of patches.
As shown in Fig.~\ref{fig:framework}, the pretraining stage involves two structurally identical networks, i.e., a student network $\phi_\mathcal{S}=f_\mathcal{S}\circ g_\mathcal{S}$ and a teacher network $\phi_\mathcal{T}=f_\mathcal{T}\circ g_\mathcal{T}$, with $f_\mathcal{S}$ and $f_\mathcal{T}$ being their respective encoders, and $g_\mathcal{S}$ and $g_\mathcal{T}$ being their respective last linear layers.
Given a sample $x$ in $D_{base}$, a set of patches $\{\hat{x}_i|i=1,2,...,n_p\}$ can be obtained by random cropping (with resize and flip).
We reserve the first $l$ ($0<l<n_p$) elements $\{\hat{x}_1,\ldots,\hat{x}_l\}$ for normal hard label supervision using the cross-entropy loss:
\begin{equation}
    \mathcal{L}_{CE}=-\mathbf{y}^\top\log{(\sigma(\frac{1}{l}\sum_{i=1}^l\phi_\mathcal{S}(\hat{x}_i)))},
\end{equation}
where $\sigma$ denotes the softmax function and $\mathbf{y}$ is the label of $x$ which is a one-hot vector.
The remaining $n_p-l$ patches $\{\hat{x}_{l+1},\ldots,\hat{x}_{n_p}\}$ are used for soft label supervision, where we construct a momentum updated \cite{mean_teacher} teacher network $\phi_\mathcal{T}$ to generate soft labels.
Specifically, denoting the parameters of $\phi_\mathcal{T}$ as $\theta_\mathcal{T}$ and those of $\phi_\mathcal{S}$ as $\theta_\mathcal{S}$, for the $i$-th iteration, $\theta_\mathcal{T}$ is updated by:
\begin{equation}
    \theta_\mathcal{T}^i\gets m\theta_\mathcal{T}^{i-1}+(1-m)\theta_\mathcal{S}^i,
\end{equation}
where $m\in[0,1)$ is a momentum coefficient.
As an exponential moving average of the student, the teacher evolves more smoothly, which ensures the stability of the generated soft labels \cite{moco,bootstrap}.
To align the output of $\phi_\mathcal{S}$ to that of $\phi_\mathcal{T}$, we propose a UniCon KL-Divergence as described below.

\subsection{UniCon KL-Divergence}

\begin{figure}[tb]
    \centering
    \includegraphics[width=\linewidth]{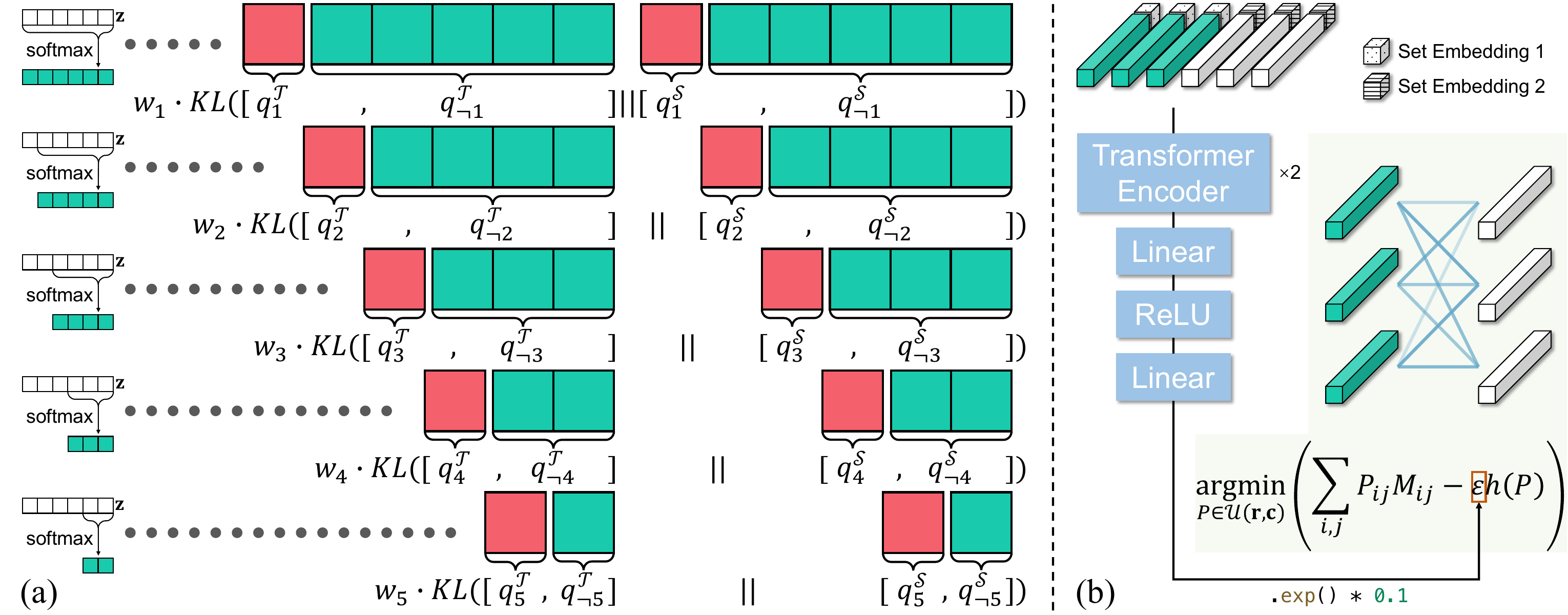}
    \caption{Illustration of (a) the continuous binary classification process corresponding to the reformulation of KL-Divergence, and (b) the proposed Adaptive Metric formulating the measure process as an OT problem. To handle various set compositions, the adjustment coefficient of an entropy regularization is tuned by a Modulate Module.}
    \label{fig:cbc_am}
\end{figure}

Denoting the network output for a patch as $\mathbf{z}=[z_1,z_2,...,z_{n_c}]\in\mathbb{R}^{n_c}$ where $z_i$ represents the logit of the $i$-th base class, the $n_c$-classification probabilities $\mathbf{p}=[p_1,p_2,...,p_{n_c}]\in\mathbb{R}^{n_c}$ can be defined where the probability of the patch belonging to class $i$ is given by:
\begin{equation}
    p_i=\frac{\exp{(z_i)}}{\sum\nolimits_{j=1}^{n_c}\exp{(z_j)}}.
\end{equation}
As a common choice to measure two probability distributions, KL-Divergence is often used for soft label supervision \cite{distill_hinton,pmlr-v80-furlanello18a,self_distill_label_smoothing}.
However, it is not suitable for learning few-shot encoders, which we elaborate on by decomposing it.

\textbf{Decomposition of KL-Divergence.}
To analyze KL-Divergence in terms of the probabilities associated with each base class, the foundation of soft labels' ability to describe patches, we consider a process shown in Fig.~\ref{fig:cbc_am}~(a).
This is a process of continuous binary classification where each time we only focus on whether the input belongs to a certain class or to the remaining classes.
The probabilities of the $i$-th binary classification $\mathbf{b}_i=[q_i,q_{\neg i}]$ can be obtained by:
\begin{equation}
    q_i=\frac{\exp{(z_i)}}{\sum\nolimits_{j=i}^{n_c}\exp{(z_j)}},\quad q_{\neg i}=\frac{\sum\nolimits_{k=i+1}^{n_c}\exp{(z_k)}}{\sum\nolimits_{j=i}^{n_c}\exp{(z_j)}}.
\end{equation}
Note that this is a process without replacement, i.e., the calculation of $\mathbf{b}_i$ only involves the logits of class $i$-$n_c$.
Converting the multivariate distribution into a series of bivariate distributions, this decomposition helps us to investigate the probabilities for distinguishing each base class, leading to the following result (the detailed proof is presented in the appendix).
\begin{theorem}\label{theorem:theorem1}
    Marking the variables calculated using the teacher output $\mathbf{z}^\mathcal{T}$ and student output $\mathbf{z}^\mathcal{S}$ with the superscripts $\mathcal{T}$ and $\mathcal{S}$, respectively, the classical KL-Divergence for soft label supervision can be reformulated as:
    \begin{equation}\label{eq:kl_decompose}
        KL(\mathbf{p}^\mathcal{T}||\mathbf{p}^\mathcal{S})=\sum\limits_{i=1}^{n_c-1}w_i\cdot KL(\mathbf{b}_i^\mathcal{T}||\mathbf{b}_i^\mathcal{S}),\quad\text{where}~w_i=\sum_{k=i}^{n_c} p_k^\mathcal{T}.
    \end{equation}
\end{theorem}

\textbf{Improper weighting scheme.}
Theorem~\ref{theorem:theorem1} demonstrates that, for KL-Divergence, the problem of measuring two classification probability distributions can be decomposed into measuring $\mathbf{b}_i$ constantly.
By giving $w_i$, it also indicates how KL-Divergence weights the measure of $\mathbf{b}_i$.
Since the continuous binary classification is a process without replacement, the number of remaining classes to be considered (class cardinality) differs for different $i$. Therefore, we consider a comparable form that normalizes $w_i$ with the class cardinality $n_c-i+1$:
\begin{equation}
    \tilde{w}_i=\frac{1}{n_c-i+1}\sum_{k=i}^{n_c} p_k^\mathcal{T}.
\end{equation}
According to $\tilde{w}_i$, the less similar the teacher thinks the input is to class $i$-$n_c$, the less important the alignment of $\mathbf{b}_i$.
This weighting scheme is consistent with the prior of normal intra-class classification tasks, i.e., the input must belong to a certain base class.
In this case, it is reasonable to stress the measure of $\mathbf{b}_i$ if the teacher thinks the input belongs to class $i$-$n_c$ or downplay it if otherwise.
However, in the context of few-shot classification, the input does not belong to any base class, and the measure of different $\mathbf{b}_i$ should be equally important as each base class prototype is equal in serving as the manifold base to represent image features \cite{causal}.

\textbf{Smoother weighting scheme for uniform base class contributions.}
Noticing that the weighting scheme can be smoothed by smoothing $\mathbf{p}^\mathcal{T}$, we introduce a temperature coefficient $T$ to alter the distribution of $\mathbf{p}^\mathcal{T}$ inspired by its use for the same purpose in various fields, e.g., contrastive learning \cite{moco} and knowledge distillation \cite{distill_hinton}.
Furthermore, we discover that the difference between the weights of two different binary classifications $\tilde{w}_\alpha(T)$ and $\tilde{w}_\beta(T)$ ($\alpha\neq\beta$) vanishes with an extremely high temperature:
\begin{equation}
    \lim_{T\to\infty}|\tilde{w}_\alpha(T)-\tilde{w}_\beta(T)|
    = \lim_{T\to\infty}\bigg|\frac{\sum\nolimits_{k_1=\alpha}^{n_c}\exp{(z_{k_1}^\mathcal{T}/T)}}{(n_c-\alpha+1)\sum\nolimits_{j=1}^{n_c}\exp{(z_j^\mathcal{T}/T)}}
    -\frac{\sum\nolimits_{k_2=\beta}^{n_c}\exp{(z_{k_2}^\mathcal{T}/T)}}{(n_c-\beta+1)\sum\nolimits_{j=1}^{n_c}\exp{(z_j^\mathcal{T}/T)}}\bigg|=0,
\end{equation}
according to which we derive a weighting scheme emphasizing the uniform contribution of different base classes for learning few-shot encoders:
\begin{equation}
    w_i^\prime\coloneqq\lim_{T\to\infty}\frac{\sum\nolimits_{k=i}^{n_c}\exp{(z_k^\mathcal{T}/T)}}{\sum\nolimits_{j=1}^{n_c}\exp{(z_j^\mathcal{T}/T)}}=\frac{n_c-i+1}{n_c}.
\end{equation}
With a more rational weighting scheme, we define UniCon KL-Divergence that is used to compute the loss for soft labels $\mathcal{L}_{UKD}$:
\begin{gather}
    UKD(\mathbf{p}^\mathcal{T}||\mathbf{p}^\mathcal{S})\coloneqq\sum\limits_{i=1}^{n_c-1}w_i^\prime\cdot KL(\mathbf{b}_i^\mathcal{T}||\mathbf{b}_i^\mathcal{S}), \\
    \mathcal{L}_{UKD}=\frac{1}{n_p-l}\sum\limits_{i=l+1}^{n_p}UKD(\sigma(\phi_T(\hat{x}_i))||\sigma(\phi_S(\hat{x}_i))).
\end{gather}
And with a weight $\lambda$, $\mathcal{L}_{UKD}$ is combined with $\mathcal{L}_{CE}$ to form the total loss for pretraining:
\begin{equation}
    \mathcal{L}_{total}=\mathcal{L}_{CE}+\lambda\cdot\mathcal{L}_{UKD}.
\end{equation}

\section{Adaptive Metric}

After pretraining, $f_\mathcal{T}$ will be used as the feature extractor for further meta-training as illustrated in Fig.~\ref{fig:framework}, where we propose Adaptive Metric for classification.

\subsection{The Sinkhorn Distance for Few-Shot Classification}

Our Adaptive Metric is based on OT distances that measure two sets by considering a hypothetical process of transporting goods from nodes of one set (suppliers) to nodes of the other set (demanders).
Given the weight vectors $\mathbf{r},\mathbf{c}\in\mathbb{R}_+^d$ ($\mathbf{r}^\top\mathbf{1}_d=\mathbf{c}^\top\mathbf{1}_d=1$) where each element represents the total supply (demand) goods of a node, and the cost per unit $M_{ij}$ for transporting from supplier $i$ to demander $j$, the goal is to find a transportation plan with the lowest total cost from a set of valid plans $\mathcal{U}(\mathbf{r},\mathbf{c})=\{P\in\mathbb{R}_+^{n\times n}|P\mathbf{1}_n=\mathbf{r},P^\top\mathbf{1}_n=\mathbf{c}\}$.

In order to endow the algorithm with the ability to properly utilize similar local relations, instead of solving $\mathop{\arg\min}\nolimits_{P\in\mathcal{U}(\mathbf{r},\mathbf{c})}\sum\nolimits_{i,j}P_{ij}M_{ij}$ directly like EMD \cite{1997The}, we formulate the set measure problem as optimizing an entropy-regularized \cite{sinkhorn} OT problem to encourage smoother transport matrices:
\begin{equation}\label{eq:sinkhorn}
    P^\varepsilon=\mathop{\arg\min}\limits_{P\in\mathcal{U}(\mathbf{r},\mathbf{c})}\Bigl(\sum\limits_{i,j}P_{ij}M_{ij}-\varepsilon h(P)\Bigr),
\end{equation}
where $\varepsilon\in(0,\infty)$ serves as an adjustment coefficient of the regularization term, and $h(P)=-\sum\nolimits_{i,j}P_{ij}\log{P_{ij}}$ is the information entropy of $P$, which reflects the smoothness of $P$, the higher the entropy, the smoother the solved matrix.

Given two local feature sets to be measured, i.e., $R(x_i^q;f)$ and $\overline{R_j}$, we define the weight of each feature with its cosine similarity to the mean of the other set, along with a softmax function to convert it to a probability distribution:
\begin{gather}
    r_i=\frac{\exp{(\hat{r}_i)}}{\sum_{j=1}^n\exp{(\hat{r}_j)}},\quad\text{where}~\hat{r}_i\coloneqq\frac{\mathbf{u}_i^\top\cdot\frac{1}{n}\sum_{j=1}^n\mathbf{v}_j}{\Vert\mathbf{u}_i\Vert\cdot\Vert\frac{1}{n}\sum_{j=1}^n\mathbf{v}_j\Vert}, \\
    c_i=\frac{\exp{(\hat{c}_i)}}{\sum_{j=1}^n\exp{(\hat{c}_j)}},\quad\text{where}~\hat{c}_i\coloneqq\frac{\mathbf{v}_i^\top\cdot\frac{1}{n}\sum_{j=1}^n\mathbf{u}_j}{\Vert\mathbf{v}_i\Vert\cdot\Vert\frac{1}{n}\sum_{j=1}^n\mathbf{u}_j\Vert}.
\end{gather}
This stems from the intuition that local features more similar to the other set are more likely to be related to the concurrent foreground objects and, hence should be assigned greater weight.
Then, with the cost to transport a unit from node $\mathbf{u}_i$ to $\mathbf{v}_j$ defined by their cosine similarity:
\begin{equation}
    M_{ij}\coloneqq1-\frac{\mathbf{u}_i^\top\mathbf{v}_j}{\Vert\mathbf{u}_i\Vert\Vert\mathbf{v}_j\Vert},
\end{equation}
we solve the optimization problem of Eq.~\ref{eq:sinkhorn} in parallel by the Sinkhorn-Knopp algorithm \cite{sinkhorn_knopp}.
Eventually, with the solved $P^\varepsilon$, we define Adaptive Metric to compute the classification score that is used for cross entropy calculation or inference:
\begin{equation}
    S(R(x_i^q;f_\mathcal{T}),\overline{R_j})\coloneqq\sum\limits_{i=1}^n\sum\limits_{j=1}^n(1-M_{ij})P_{ij}^\varepsilon.
\end{equation}

\subsection{Modulate Module}

Furthermore, to control the smoothness of the transport matrix adaptively according to specific local feature sets, we design a Modulate Module to predict $\varepsilon$ in Eq.~\ref{eq:sinkhorn} instead of treating it as a pre-fixed hyperparameter.
By giving higher $\varepsilon$, $P^\varepsilon$ will be smoother, and as $\varepsilon$ goes to zero, it will be sparser, with the solution close to EMD.
Intuitively, the smoothness of the transport matrix should be conditioned on the relationship of the local features (similar features come with similar local patches where a smooth transport matrix is expected). Therefore, we take the extracted local features as input and construct a predictor based on the Transformer encoder \cite{transformer}, considering that its inductive bias suits the task of modeling the relationship between local features very well.
As shown in Fig.~\ref{fig:cbc_am}~(b), the input embeddings are constructed by concatenating the local feature with a $16$ dimensional learnable set embedding indicating which set the local feature is from, i.e., $R(x_i^q;f_\mathcal{T})$ or $\overline{R_j}$. Followed by an exponential function, the output serves as a scaling factor to adjust $\varepsilon$ from the default value of $0.1$.

The overall training process of our method is described in Algorithm~\ref{alg:alg1}.

\begin{algorithm}
\caption{Training process of FCAM.}\label{alg:alg1}
\begin{multicols}{2}
\begin{algorithmic}[1]
\Statex \textsc{Pretraining}
\State Warm up $\phi_\mathcal{S}$ with $\mathcal{L}_{CE}$;
\State $\theta_\mathcal{T}\gets\theta_\mathcal{S}$;
\While{$\text{epochs}$}
\While{$\text{steps}$}
\State Randomly crop $n_p$ patches $\{\hat{x}_1,\ldots,\hat{x}_{n_p}\}$ for each image $x$ in the minibatch;
\State Calculate $\mathcal{L}_{CE}$ with $\{\hat{x}_1,\ldots,\hat{x}_l\}$;
\State Calculate $\mathcal{L}_{UKD}$ with $\{\hat{x}_{l+1},\ldots,\hat{x}_{n_p}\}$;
\State $\mathcal{L}_{total}=\mathcal{L}_{CE}+\lambda\cdot\mathcal{L}_{UKD}$;
\State Update $\theta_\mathcal{S}$ with $\nabla_{\theta_\mathcal{S}}\mathcal{L}_{total}$;
\State Update $\theta_\mathcal{T}$, i.e., $\theta_\mathcal{T}^i\gets m\theta_\mathcal{T}^{i-1}+(1-m)\theta_\mathcal{S}^i$;
\EndWhile
\EndWhile
\end{algorithmic}
\columnbreak
\begin{algorithmic}[1]
\Statex \textsc{Meta-Training}
\While{not converged}
\State Construct a task, i.e., sample $D_{spt}$, $D_{qry}$ from $D_{base}$;
\State Calculate $\overline{R_j}$ \textbf{for} $j$ \textbf{in} $N$;
\For{$x_i^q$ \textbf{in} $D_{qry}$}
\State Calculate $R(x_i^q;f_\mathcal{T})$;
\State Predict $\varepsilon$ and calculate $S(R(x_i^q;f_\mathcal{T}),\overline{R_j})$ \textbf{for} $j$ \textbf{in} $N$;
\EndFor
\State Calculate cross entropy loss;
\State Optimize $f_\mathcal{T}$;
\EndWhile
\end{algorithmic}
\end{multicols}
\vspace{-1em}
\end{algorithm}

\section{Experiments}

\textbf{Datasets.} The experiments are conducted on three popular benchmarks: (1) \textbf{\textit{mini}ImageNet} \cite{matchingnet} is a subset of ImageNet \cite{imagenet} that contains 100 classes with 600 images per class. The 100 classes are divided into 64/16/20 for train/val/test respectively; (2) \textbf{\textit{tiered}ImageNet} \cite{tiered} is also a subset of ImageNet \cite{imagenet} that includes 608 classes from 34 super-classes. The super-classes are split into 20/6/8 for train/val/test respectively; (3) \textbf{CUB-200-2011} \cite{cub} contains 200 bird categories with 11,788 images, which represents a fine-grained scenario. Following the splits in \cite{FEAT}, the 200 classes are divided into 100/50/50 for train/val/test respectively.

\textbf{Backbone.} For the backbone, we employ \textit{ResNet12} as many previous works. With the dimension of the embedded features and the set embeddings being 640 and 16, respectively, we set $d_{model}=656$, $d_{feedforward}=1280$ and $n_{head}=16$ for the $2$-layer Transformer encoder in our Modulate Module.

\textbf{Training details.} In the pretraining stage, we set $n_p=4$, $l=1$ and $m=0.999$. $\mathcal{L}_{UKD}$ will not be used during early epochs to ensure the teacher has well-converged before being used to generate soft labels.
In the meta-training stage, each epoch involves $50$ iterations with a batch size of $4$. We set $n=25$, and the patches are resized to $84\times84$ before being embedded. The Modulate Module is first trained for $100$ epochs with the encoder's parameters fixed, in which the learning rate starts from $1e$-$3$ and decays by $0.1$ at epoch $60$ and $90$. Then, all the parameters will be optimized jointly for another $100$ epochs.

\subsection{Comparison with State-of-the-art Methods}

For general few-shot classification, we compare our method with the state-of-the-art methods in Table~\ref{tab:mini_tiered}. Our method outperforms the state-of-the-art methods on all the settings and even achieves higher performance than methods with bigger backbones, achieving new state-of-the-art performance.
For fine-grained few-shot classification, we compare our method with the state-of-the-art methods in Table~\ref{tab:cub}. Benefit from higher quality local features, the discriminative regions can be depicted more accurately, resulting in significant improvement against other methods, i.e., $\bf{4.41\%}$ and $\bf{2.85\%}$ for $1$-shot and $5$-shot respectively against previous state-of-the-art method \cite{random_walk}. In particular, our method even outperforms state-of-the-art transductive \cite{ECKPN,ADRGN} and cross-modal \cite{AGAM,ADRGN} methods, shedding some light on how much the poor local representations can degrade the performance in the fine-grained scenario.

\subsection{Ablation Study}

To begin with, a coarse-scale ablation is presented in Table~\ref{tab:ablation}. The baseline follows the traditional pretraining paradigm that uses only $\mathcal{L}_{CE}$ for supervision and employs EMD as the metric. With both Feature Calibration and Adaptive Metric outperforming the baseline and achieving optimal results when used together, their respective effectiveness can be validated. Furthermore, we conduct a more detailed analysis below.

\begin{table}[htb]
    \caption{Comparison to the state-of-the-art methods on \textit{mini}ImageNet and \textit{tiered}ImageNet, ordered chronologically. Average $5$-way $1$-shot and $5$-way $5$-shot accuracy ($\%$) with $95\%$ confidence intervals.}
    \label{tab:mini_tiered}
    \centering\resizebox{\textwidth}{!}{
    \begin{threeparttable}
    \begin{tabular}{lccccc}
        \toprule
        \multirow{2}{*}{Method} & \multirow{2}{*}{Backbone} & \multicolumn{2}{c}{\textit{mini}ImageNet} & \multicolumn{2}{c}{\textit{tiered}ImageNet} \\ \cmidrule{3-6}
        & & $1$-shot & $5$-shot & $1$-shot & $5$-shot \\
        \midrule
        MatchNet\tnote{$\dagger$}~~\cite{matchingnet} & \textit{ResNet12} & $63.08\pm0.80$ & $75.99\pm0.60$ & $68.50\pm0.92$ & $80.60\pm0.71$ \\
        ProtoNet\tnote{$\dagger$}~~\cite{protonet} & \textit{ResNet12} & $60.37\pm0.83$ & $78.02\pm0.57$ & $65.65\pm0.92$ & $83.40\pm0.65$ \\
        TADAM \cite{TADAM} & \textit{ResNet12} & $58.50\pm0.30$ & $76.70\pm0.30$ & - & - \\
        FEAT \cite{FEAT} & \textit{ResNet12} &  $66.78\pm0.20$ &  $ 82.05\pm0.14$ & $70.80\pm0.23$ & $84.79\pm0.16$ \\
        DeepEMD \cite{EMD} & \textit{ResNet12} & $65.91\pm0.82$ & $82.41\pm0.56$ & $71.16\pm0.87$ & $86.03\pm0.58$ \\
        Meta-Baseline \cite{meta_baseline} & \textit{ResNet12} & $63.17\pm0.23$ & $ 79.26\pm0.17$ & $68.62\pm0.27$ & $ 83.74\pm0.18$ \\
        FRN \cite{FRN} & \textit{ResNet12} & $66.45\pm0.19$ & $82.83\pm0.13$ & $72.06\pm0.22$ & $86.89\pm0.14$ \\
        PAL \cite{PAL} & \textit{ResNet12} & $\underline{69.37\pm0.64}$ & $84.40\pm0.44$ & $72.25\pm0.72$ & $86.95\pm0.47$ \\
        MCL \cite{random_walk} & \textit{ResNet12} & $69.31\pm0.20$ & $\underline{85.11\pm0.20}$ & $73.62\pm0.20$ & $86.29\pm0.20$ \\
        DeepEMD v2 \cite{EMD2} & \textit{ResNet12} & $68.77\pm0.29$ & $84.13\pm0.53$ & $\underline{74.29\pm0.32}$ & $\underline{87.08\pm0.60}$ \\
        FADS \cite{few_shot_23} & \textit{ResNet12} & $66.73\pm0.88$ & $83.51\pm0.51$ & $74.12\pm0.74$ & $86.56\pm0.46$ \\
        \midrule
        Centroid Alignment\tnote{$\ddagger$}~~\cite{centroid_alignment} & \textit{WRN-28-10} & $65.92\pm0.60$ & $82.85\pm0.55$ & $74.40\pm0.68$ & $86.61\pm0.59$ \\
        Oblique Manifold\tnote{$\ddagger$}~~\cite{oblique_manifold} & \textit{ResNet18} & $63.98\pm0.29$ & $82.47\pm0.44$ & $70.50\pm0.31$ & $86.71\pm0.49$ \\
        FewTURE\tnote{$\ddagger$}~~\cite{fewture} & \textit{ViT-Small} & $68.02\pm0.88$ & $84.51\pm0.53$ & $72.96\pm0.92$ & $86.43\pm0.67$ \\
        \midrule
        FCAM (ours) & \textit{ResNet12} & $\bf{70.47\pm0.28}$ & $\bf{85.71\pm0.52}$ & $\bf{75.02\pm0.31}$ & $\bf{88.41\pm0.59}$ \\
        \bottomrule
    \end{tabular}
    \begin{tablenotes}
        \footnotesize
        \item[$\dagger$] results are reported in \cite{EMD2}.
        \item[$\ddagger$] methods with bigger backbones.
        \item The second best results are \underline{underlined}.
    \end{tablenotes}
    \end{threeparttable}}
\end{table}

\begin{table}[htb]
\centering
\begin{minipage}[t]{0.55\linewidth}
    \caption{Comparison to the state-of-the-art methods on CUB-200-2011, ordered chronologically. Average $5$-way $1$-shot and $5$-way $5$-shot accuracy ($\%$) with $95\%$ confidence intervals.}
    \label{tab:cub}
    \centering\resizebox{\textwidth}{!}{
    \begin{threeparttable}
    \begin{tabular}{lccc}
        \toprule
        \multirow{2}{*}{Method} & \multirow{2}{*}{Backbone} & \multicolumn{2}{c}{CUB-200-2011} \\ \cmidrule{3-4}
        & & $1$-shot & $5$-shot \\
        \midrule
        MatchNet\tnote{$\dagger$}~~\cite{matchingnet} & \textit{ResNet12} & $71.87\pm0.85$ & $85.08\pm0.57$ \\
        ProtoNet\tnote{$\dagger$}~~\cite{protonet} & \textit{ResNet12} & $66.09\pm0.92$ & $82.50\pm0.58$ \\
        DeepEMD \cite{EMD} & \textit{ResNet12} & $75.65\pm0.83$ & $88.69\pm0.50$ \\
        FRN\tnote{$\sharp$}~~\cite{FRN} & \textit{ResNet12} & $78.86\pm0.28$ & $\underline{90.48\pm0.16}$ \\
        MCL\tnote{$\sharp$}~~\cite{random_walk} & \textit{ResNet12} & $\underline{79.39\pm0.29}$ & $\underline{90.48\pm0.49}$ \\
        DeepEMD v2 \cite{EMD2} & \textit{ResNet12} & $79.27\pm0.29$ & $89.80\pm0.51$ \\
        \midrule
        Centroid Alignment\tnote{$\ddagger$}~~\cite{centroid_alignment} & \textit{ResNet18} & $74.22\pm1.09$ & $88.65\pm0.55$ \\
        Oblique Manifold\tnote{$\ddagger$}~~\cite{oblique_manifold} & \textit{ResNet18} & $78.24\pm-$ & $92.15\pm-$ \\
        ECKPN\tnote{$\flat$}~~\cite{ECKPN} & \textit{ResNet12} & $77.43\pm0.54$ & $92.21\pm0.41$ \\
        AGAM\tnote{$\natural$}~~\cite{AGAM} & \textit{ResNet12} & $79.58\pm0.25$ & $87.17\pm0.23$ \\
        ADRGN\tnote{$\flat\natural$}~~~\cite{ADRGN} & \textit{ResNet12} & $82.32\pm0.51$ & $92.97\pm0.35$ \\
        \midrule
        FCAM (ours) & \textit{ResNet12} & $\bf{82.89\pm0.27}$ & $\bf{93.06\pm0.39}$ \\
        \bottomrule
    \end{tabular}
    \begin{tablenotes}
        \footnotesize
        \item[$\dagger$] results are reported in \cite{EMD2}.
        \item[$\ddagger$] methods with bigger backbones.
        \item[$\sharp$] reproduced using the data split we use.
        \item[$\flat$] transductive methods.
        \item[$\natural$] methods that use attribute information.
        \item The second best results are \underline{underlined}.
    \end{tablenotes}
    \end{threeparttable}}
\end{minipage}
\hfill
\begin{minipage}[t]{0.44\linewidth}
    \caption{Ablation of Feature Calibration and Adaptive Metric. The experiments are conducted with \textit{ResNet12} on \textit{mini}ImageNet.}
    \label{tab:ablation}
    \centering\resizebox{\textwidth}{!}{
    \begin{tabular}{cccc}
        \toprule
        Feature & Adaptive & \multirow{2}{*}{$1$-shot} & \multirow{2}{*}{$5$-shot} \\ Calibration & Metric & & \\
        \midrule
        & & $67.57\pm0.29$ & $83.39\pm0.54$ \\
        \checkmark & & $69.54\pm0.29$ & $85.28\pm0.52$ \\
        & \checkmark & $69.01\pm0.28$ & $84.41\pm0.53$ \\
        \checkmark & \checkmark & $\bf{70.47\pm0.28}$ & $\bf{85.71\pm0.52}$ \\
        \bottomrule
    \end{tabular}}
    \vspace{0.7em}
    \caption{Comparison of using classical and UniCon KL-Divergence for calibration (top), and the results of whether using Modulate Module to adjust $\varepsilon$ (bottom).}
    \label{tab:ablation_UKD_RM}
    \centering\resizebox{\linewidth}{!}{
    \begin{tabular}{lccc}
        \toprule
        Setting & $1$-shot & $5$-shot \\
        \midrule
        Classical KL-Divergence & $69.94\pm0.28$ & $84.79\pm0.53$ \\
        UniCon KL-Divergence & $\bf{70.47\pm0.28}$ & $\bf{85.71\pm0.52}$ \\
        \midrule
        \midrule
        w/o Modulate Module & $69.69\pm0.28$ & $85.23\pm0.52$ \\
        w/ Modulate Module & $\bf{70.47\pm0.28}$ & $\bf{85.71\pm0.52}$ \\
        \bottomrule
    \end{tabular}}
\end{minipage}
\end{table}

\begin{wrapfigure}{t}{0.55\linewidth}
    \centering
    \includegraphics[width=\linewidth]{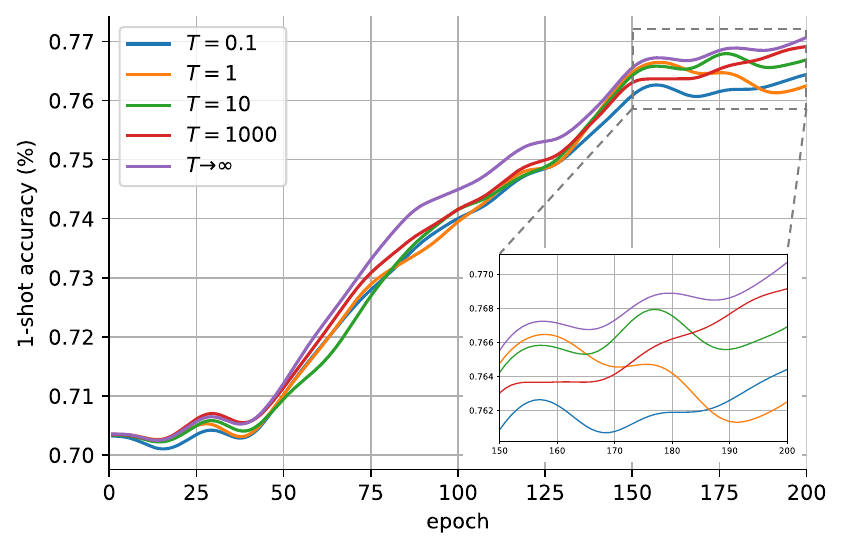}\vspace{-1em}
    \caption{Gaussian smoothed $1$-shot test accuracy curves on CUB-200-2011 during feature calibration, with different temperatures to adjust the weighting scheme of the classical KL-Divergence. The results of the same $1000$ tasks are averaged for each data point.}
    \label{fig:temp}\vspace{1.7em}
    \centering
    \includegraphics[width=\linewidth]{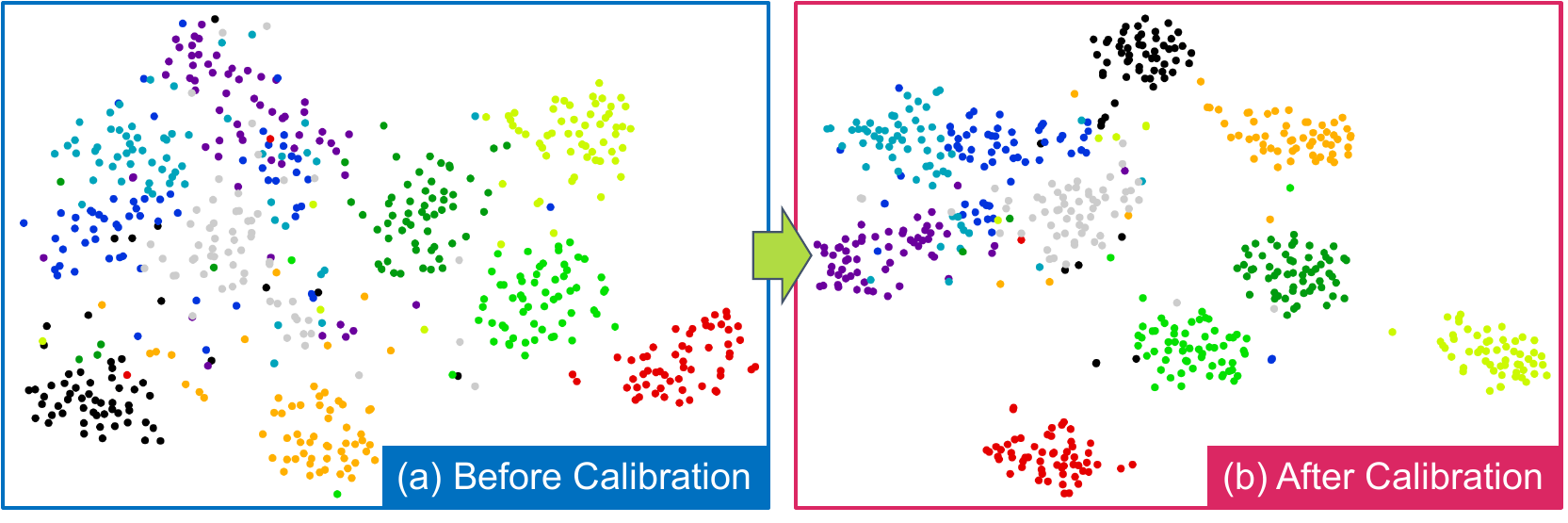}\vspace{-1em}
    \caption{Visualization \cite{tsne} of novel class samples embedded by encoders trained (a) without and (b) with Feature Calibration.}
    \label{fig:tsne}\vspace{-3em}
\end{wrapfigure}

\textbf{Feature Calibration improves novel-class generalization.}
To demonstrate that Feature Calibration improves novel-class generalization, we visualize the $1$-shot test accuracy change during feature calibration in Fig.~\ref{fig:temp}.
We first pre-train the network to its highest validation accuracy with only $\mathcal{L}_{CE}$ to ensure the quality of the teacher and exclude the influence of hard label supervision on accuracy improvement during calibration.
We observe a continuous improvement in test accuracy during calibration. In the case of our method ($T\to\infty$), the $1$-shot accuracy is boosted from $70.20\%$ to $77.41\%$, demonstrating the effectiveness of Feature Calibration in improving novel-class generalization and suggesting how severe the power of local representations is limited. In addition, the feature distributions visualized in Fig.~\ref{fig:tsne} also illustrate that Feature Calibration results in better clusters for novel classes.

\textbf{UniCon KL-Divergence is more suitable for Feature Calibration.}
We compare different temperature settings in Fig.~\ref{fig:temp}. It can be seen that the temperature, i.e., the weighting scheme, affects the process of Feature Calibration.
A general trend that better test accuracy comes with higher temperature can be observed, and the setting corresponding to our UniCon KL-Divergence, i.e., $T\to\infty$, constantly outperforms other settings.
Furthermore, UniCon KL-Divergence yields better final performance than classical KL-Divergence as shown in Table~\ref{tab:ablation_UKD_RM}. Both the above experiments demonstrate the importance of a smoother weighting scheme in Feature Calibration.

\begin{figure}[tb]
    \centering
    \includegraphics[width=\linewidth]{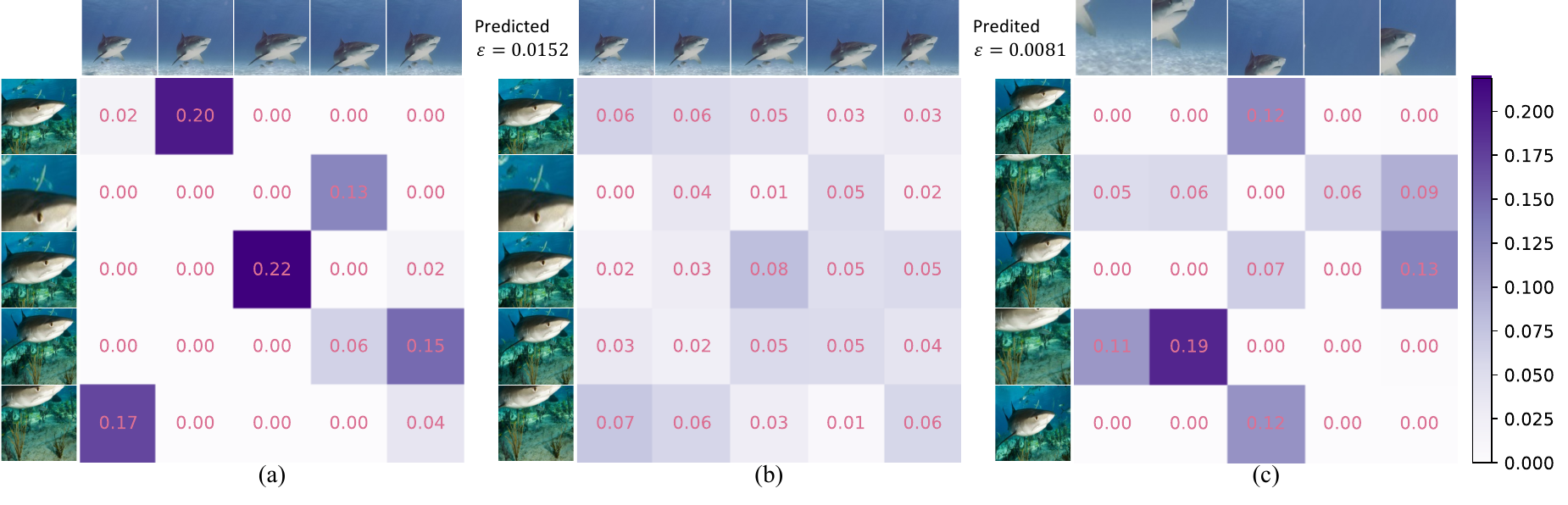}\vspace{-2em}
    \caption{Visualization of solved transport matrices. Results of (a) EMD and (b) Adaptive Metric for sets consisting of similar local features, and the result of (c) Adaptive Metric for sets consisting of dissimilar local features.}
    \label{fig:match_vis}
\end{figure}

\textbf{Entropic term handles sets consisting of similar nodes.}
For sets consisting of similar local features, the transport matrix solved by EMD (Fig.~\ref{fig:match_vis}~(a)) is very sparse, ignoring a lot of similar local relations. In contrast, Adaptive Metric (Fig.~\ref{fig:match_vis}~(b)) is able to generate a smoother transport matrix due to the entropic regularization, which enables a comprehensive utilization of similar local relations and reduces the dependency on a few of them by allowing ``one-to-many'' matching.

\textbf{Modulate Module brings adaptability.}
For sets consisting of similar local features, Modulate Module predicts a relatively larger $\varepsilon$, resulting in a smoother transport matrix (Fig.~\ref{fig:match_vis}~(b)). For sets consisting of dissimilar local features, a relatively smaller $\varepsilon$ is produced, making the transport matrix moderately sparse (Fig.~\ref{fig:match_vis}~(c)). Quantitative results of whether using Modulate Module to adjust $\varepsilon$ is also presented in Table~\ref{tab:ablation_UKD_RM}. Compared to a fixed default value, it introduces adaptability into the measure process, helping achieve better performance.

\subsection{Cross-Domain Experiments}

For the cross-domain setting which poses a greater challenge for novel-class generalization, we perform an experiment where models are trained on \textit{mini}Imagenet and evaluated on CUB-200-2011.
This setting allows us to better evaluate the model's ability to handle novel classes with significant domain differences from the base classes, due to the large domain gap.
As shown in Table~\ref{tab:cross_domain}, our method outperforms the previous state of the art, demonstrating its superiority in improving novel-class generalization.

\begin{table}[tb]
\centering
\begin{minipage}[t]{0.45\linewidth}
    \caption{Cross-domain experiments following the setting of \cite{chen2019closerfewshot} (\textit{mini}ImageNet$\rightarrow$CUB). Average $5$-way $1$-shot and $5$-way $5$-shot accuracy ($\%$) with $95\%$ confidence intervals.}
    \label{tab:cross_domain}
    \centering\resizebox{\linewidth}{!}{
    \begin{threeparttable}
    \begin{tabular}{lcc}
        \toprule
        Method & $1$-shot & $5$-shot \\
        \midrule
        ProtoNet\tnote{$\dagger$}~~\cite{protonet} & $50.01\pm0.82$ & $72.02\pm0.67$ \\
        MatchNet\tnote{$\dagger$}~~\cite{matchingnet} & $51.65\pm0.84$ & $69.14\pm0.72$ \\
        \textit{cosine} classifier \cite{chen2019closerfewshot} & $44.17\pm0.78$ & $69.01\pm0.74$ \\
        \textit{linear} classifier \cite{chen2019closerfewshot} & $50.37\pm0.79$ & $73.30\pm0.69$ \\
        KNN \cite{DN4} & $50.84\pm0.81$ & $71.25\pm0.69$ \\
        DeepEMD v2 \cite{EMD2} & $\underline{54.24\pm0.86}$ & $\underline{78.86\pm0.65}$ \\
        FCAM (ours) & $\bf{58.20\pm0.30}$ & $\bf{80.92\pm0.65}$ \\
        \bottomrule
    \end{tabular}
    \begin{tablenotes}
        \footnotesize
        \item[$\dagger$] results are reported in \cite{EMD2}.
        \item The second best results are \underline{underlined}.
    \end{tablenotes}
    \end{threeparttable}}
\end{minipage}
\hfill
\begin{minipage}[t]{0.54\linewidth}
    \caption{Results of global-representation-based FSC methods on \textit{mini}ImageNet, w/o and w/ Feature Calibration (FC). Average $5$-way $1$-shot and $5$-way $5$-shot accuracy ($\%$) with $95\%$ confidence intervals.}
    \label{tab:fc_global}
    \centering\resizebox{\linewidth}{!}{
    \begin{tabular}{lccc}
        \toprule
        Method & Setting & $1$-shot & $5$-shot \\
        \midrule
        \multirow{2}{*}{\textit{cosine} classifier \cite{chen2019closerfewshot}} & w/o FC & $61.31\pm0.20$ & $77.73\pm0.21$ \\
        & w/ FC & $\bf{64.92\pm0.20}$ & $\bf{80.51\pm0.21}$ \\
        \midrule
        \multirow{2}{*}{\textit{linear} classifier \cite{chen2019closerfewshot}} & w/o FC & $55.74\pm0.20$ & $78.89\pm0.21$ \\
        & w/ FC & $\bf{59.35\pm0.20}$ & $\bf{81.46\pm0.20}$ \\
        \midrule
        \multirow{2}{*}{Classifier-Baseline \cite{meta_baseline}} & w/o FC & $60.67\pm0.21$ & $78.53\pm0.21$ \\
        & w/ FC & $\bf{64.33\pm0.21}$ & $\bf{81.01\pm0.21}$ \\
        \midrule
        \multirow{2}{*}{Meta-Baseline \cite{meta_baseline}} & w/o FC & $63.62\pm0.21$ & $80.25\pm0.20$ \\
        & w/ FC & $\bf{64.90\pm0.21}$ & $\bf{81.04\pm0.21}$ \\
        \bottomrule
    \end{tabular}}
\end{minipage}
\end{table}

\subsection{Feature Calibration for Global-Representation-based FSC}

Although Feature Calibration is proposed for improving local representations, it also benefits methods based on global representations as demonstrated in Table~\ref{tab:fc_global}.
Feature Calibration boosts the performance of these methods significantly due to its ability to leverage the class-level diversity provided by random cropping.

\section{Conclusion}

In this paper, we presented a novel FCAM method for few-shot classification to harness the power of local representations in improving novel-class generalization.
It improves the few-shot encoder by calibrating it towards the test scenario and handles various set compositions of local feature sets adaptively.
Our method achieves new state-of-the-art performance on multiple datasets.
To further enhance FCAM, we will seek better strategies for constructing local feature sets because a notable limitation of FCAM is the reliance of the performance on the patch number, which brings considerable computational cost.

\bibliographystyle{unsrt}
\bibliography{references}

\appendix

\renewcommand{\thefigure}{\Alph{figure}}
\renewcommand{\thetable}{\Alph{table}}
\setcounter{figure}{0}
\setcounter{table}{0}

\vspace{3em}\centerline{\large \bfseries \scshape Appendix}

The appendix is organized as follows:
\begin{itemize}
    \item Sec.~\ref{sec:proof} presents the proof of Theorem~\ref{theorem:theorem1};
    \item Sec.~\ref{sec:experimental_setup} describes our experimental setup in detail;
    \item Sec.~\ref{sec:addtional_results} shows some additional experimental results, including some analysis on computational time (Sec.~\ref{subsec:time}) and more visualized transport matrices (Sec.~\ref{subsec:match}) as a supplement to Fig.~\ref{fig:match_vis}.
\end{itemize}

\section{Proof of Theorem~\ref{theorem:theorem1}}\label{sec:proof}

According to the definition of $q_i$ and $w_i$, $q_i=\frac{p_i}{\sum\nolimits_{k=i}^{n_c}p_k}$, hence $p_i^\mathcal{T} = w_i \cdot q_i^\mathcal{T}$. Therefore, we have:
\begin{gather}
    KL(\mathbf{p}^\mathcal{T}||\mathbf{p}^\mathcal{S})
    = \sum_{i=1}^{n_c} p_i^\mathcal{T} \log\frac{p_i^\mathcal{T}}{ p_i^\mathcal{S}}
    = \sum_{i=1}^{n_c} p_i^\mathcal{T} \log\frac{q_i^\mathcal{T}}{q_i^\mathcal{S}} + \sum_{i=1}^{n_c} p_i^\mathcal{T} \log\frac{\sum_{k=i}^{n_c} p_k^\mathcal{T}}{\sum_{k=i}^{n_c} p_k^\mathcal{S}}, \\
    \sum_{i=1}^{n_c} p_i^\mathcal{T} \log\frac{q_i^\mathcal{T}}{q_i^\mathcal{S}}
    = \sum_{i=1}^{n_c} w_i \cdot q_i^\mathcal{T}\log\frac{q_i^\mathcal{T}}{q_i^\mathcal{S}}.
\end{gather}
And according to $q_{\neg i}=1-q_i=\frac{\sum\nolimits_{k=i+1}^{n_c}p_k}{\sum\nolimits_{k=i}^{n_c}p_k}$, we have $\sum_{k=i+1}^{n_c} p_k^\mathcal{T} = w_i\cdot q_{\neg i}^\mathcal{T}$, and:
\begin{equation}
    \sum_{k=i}^{n_c} p_k=q_{\neg(i-1)} \cdot \sum_{k=i-1}^{n_c} p_k=(\prod_{k=1}^{i-1} q_{\neg k}) \cdot (\sum_{k=1}^{n_c} p_k)=\prod_{k=1}^{i-1} q_{\neg k}. \label{eqn:elabel4}
\end{equation}
From the above equation, it can be concluded that:
\begin{equation}
\begin{split}
    \sum_{i=1}^{n_c} p_i^\mathcal{T} \log\frac{\sum_{k=i}^{n_c} p_k^\mathcal{T}}{\sum_{k=i}^{n_c} p_k^\mathcal{S}}
    =& \sum_{i=1}^{n_c} p_i^\mathcal{T}\log{(\prod_{k=1}^{i-1}\frac{ q_{\neg k}^\mathcal{T}}{ q_{\neg k}^\mathcal{S}})}
    = \sum_{i=1}^{n_c} \sum_{k=1}^{i-1} p_i^\mathcal{T}\log\frac{ q_{\neg k}^\mathcal{T}}{ q_{\neg k}^\mathcal{S}} \\
    =& \sum_{k=1}^{n_c-1} \sum_{i=k+1}^{n_c} p_i^\mathcal{T}\log\frac{ q_{\neg k}^\mathcal{T}}{ q_{\neg k}^\mathcal{S}}
    = \sum_{k=1}^{n_c-1} w_k \cdot q_{\neg k}^\mathcal{T}\log\frac{ q_{\neg k}^\mathcal{T}}{ q_{\neg k}^\mathcal{S}}.
\end{split}
\end{equation}
Therefore, the KL-Divergence can be reformulated as follows:
\begin{equation}
\begin{split}
    &KL(\mathbf{p}^\mathcal{T}||\mathbf{p}^\mathcal{S})
    = \sum_{i=1}^{n_c} w_i \cdot q_i^\mathcal{T}\log\frac{q_i^\mathcal{T}}{q_i^\mathcal{S}}
    + \sum_{k=1}^{n_c-1} w_k \cdot q_{\neg k}^\mathcal{T}\log\frac{ q_{\neg k}^\mathcal{T}}{ q_{\neg k}^\mathcal{S}} \\
    =& \sum_{i=1}^{n_c-1}w_i\cdot(q_i^\mathcal{T}\log\frac{q_i^\mathcal{T}}{q_i^\mathcal{S}}
    + q_{\neg i}^\mathcal{T}\log\frac{ q_{\neg i}^\mathcal{T}}{ q_{\neg i}^\mathcal{S}})
    = \sum_{i=1}^{n_c-1} w_i\cdot KL(\mathbf{b}_i^\mathcal{T}||\mathbf{b}_i^\mathcal{S}).
\end{split}
\end{equation}

\section{Detailed Experimental Setup}\label{sec:experimental_setup}

Following the ``pretraining + meta-training'' paradigm, the training process of our method can be divided into two stages.
For the pretraining stage, we pre-train the encoder with our proposed pretraining paradigm based on the proxy task of standard multi-classification on $D_{base}$ and select the model with the highest validation accuracy.
For the meta-training stage, each epoch involves 50 iterations with a batch size of 4. We first pre-train the Modulate Module with the parameters of the encoder fixed. Then, the parameters of both the encoder and the Modulate Module are optimized jointly.
Globally, we set $n_p=4$, $l=1$, $m=0.999$, and $n=25$. The cropped patches are resized to $84\times84$ before being embedded.
For evaluation, we randomly sample $5000$/$600$ episodes for testing and report the average accuracy with the $95\%$ confidence interval for $1$-shot/$5$-shot experiment following \cite{EMD2}.
In the following, we describe our detailed experimental setup according to different benchmarks:

\begin{itemize}
    \item[(1)] \textbf{\textit{mini}ImageNet.} The encoder is first pre-trained for $360$ epochs where the SGD optimizer with a momentum of $0.9$ and a weight decay of $5e$-$4$ is adopted. $\mathcal{L}_{UKD}$ will not be used for the first $120$ epochs to ensure the teacher has well-converged before being used. For the latter $240$ epochs, the learning rate is set to $0.01$, and $\lambda$ is set to $0.1$. Then, in an episodic manner, the Modulate Module is pre-trained for $100$ epochs with the parameters of the encoder fixed, in which the Adam optimizer with a weight decay of $5e$-$4$ is adopted. The learning rate starts from $1e$-$3$ and decays by $0.1$ at epoch $60$ and $90$. Finally, all the parameters will be optimized jointly for another $100$ epochs where the SGD optimizer with a momentum of $0.9$ and a weight decay of $5e$-$4$ is adopted. The learning rate starts from $5e$-$4$ and decays by $0.5$ every 10 epochs.
    \item[(2)] \textbf{\textit{tiered}ImageNet.} The encoder is first pre-trained for $240$ epochs where the SGD optimizer with a momentum of $0.9$ and a weight decay of $5e$-$4$ is adopted. $\mathcal{L}_{UKD}$ will not be used for the first $120$ epochs to ensure the teacher has well-converged before being used. For the latter $120$ epochs, the learning rate is set to $0.001$, and $\lambda$ is set to $0.05$. Then, in an episodic manner, the Modulate Module is pre-trained for $100$ epochs with the parameters of the encoder fixed, in which the Adam optimizer with a weight decay of $5e$-$4$ is adopted. The learning rate starts from $1e$-$3$ and decays by $0.1$ at epoch $60$ and $90$. Finally, all the parameters will be optimized jointly for another $100$ epochs where the SGD optimizer with a momentum of $0.9$ and a weight decay of $5e$-$4$ is adopted. The learning rate starts from $1e$-$4$ and decays by $0.5$ every 10 epochs.
    \item[(3)] \textbf{CUB-200-2011.} Each image is first cropped with the provided human-annotated bounding box as many previous works \cite{NIPS2017_01e9565c,random_walk,EMD2}. The encoder is first pre-trained for $360$ epochs where the SGD optimizer with a momentum of $0.9$ and a weight decay of $5e$-$4$ is adopted. $\mathcal{L}_{UKD}$ will not be used for the first $120$ epochs to ensure the teacher has well-converged before being used. For the latter $240$ epochs, the learning rate is set to $0.03$, and $\lambda$ is set to $0.5$. Then, in an episodic manner, the Modulate Module is pre-trained for $100$ epochs with the parameters of the encoder fixed, in which the Adam optimizer with a weight decay of $5e$-$4$ is adopted. The learning rate starts from $1e$-$3$ and decays by $0.1$ at epoch $60$ and $90$. Finally, all the parameters will be optimized jointly for another $100$ epochs where the SGD optimizer with a momentum of $0.9$ and a weight decay of $5e$-$4$ is adopted. The learning rate starts from $1e$-$3$ and decays by $0.5$ every 10 epochs.
\end{itemize}

\section{Additional Experimental Results}\label{sec:addtional_results}

\subsection{Analysis on Computational Time}\label{subsec:time}

Although the Modulate Module will inevitably bring additional time overhead during inference as a parameterized module, the proposed Adaptive Metric still costs less time for measuring two feature sets compared to EMD since the solution of the OT problem is accelerated. Specifically, the introduced entropic regularization makes the OT problem a strictly convex problem \cite{sinkhorn}. Thus, it can be solved by the Sinkhorn-Knopp algorithm \cite{sinkhorn_knopp} which is known to have a linear convergence \cite{FRANKLIN1989717,doi:10.1137/060659624}.
We conduct an experiment on an RTX-3090 (Linux, PyTorch 3.6) using the same $10,000$ randomly sampled episodes to compare the time cost empirically. The average time spent to process an episode is reported in Table~\ref{tab:time}.
It can be seen that our Adaptive Metric spends way less time than EMD ($26.33\%$ faster) even in the presence of a parameterized module, demonstrating its superiority in both accuracy and speed.

We also analyzed the influence of the number of patches used to represent an image as shown in Table~\ref{tab:patch_num}.
While it shows good robustness, a limitation of FCAM is also revealed. Better accuracy requires more patches, which will result in more time for embedding them and solving the optimal transport problem.
To mitigate the dependence of accuracy on the number of patches and reduce the computational cost, how to actively select most relevant patches for constructing local feature sets is an important problem to be addressed in the future.

\begin{table}[htb]
    \caption{Time spent processing an episode for methods with different metrics. 9 patches are used to represent a sample.}
    \label{tab:time}
    \centering
    \begin{tabular}{lccc}
        \toprule
        Metric & Average time per task (ms) \\
        \midrule
        EMD & 378.42 \\
        Adaptive Metric (ours) & \bf{278.78} \\
        \bottomrule
    \end{tabular}
\end{table}

\begin{table}[htb]
    \caption{The $1$-shot accuracy and the average time spent per task when representing an image with different numbers of patches.}
    \label{tab:patch_num}
    \centering
    \begin{tabular}{ccccc}
        \toprule
        \multirow{2}{*}{\# of patches} & \multicolumn{2}{c}{\textit{mini}ImageNet} & \multicolumn{2}{c}{CUB-200-2011} \\ \cmidrule{2-5} & $1$-shot acc & Time/task (ms) & $1$-shot acc & Time/task (ms) \\
        \midrule
        $9$ & $69.42\pm0.28$ & $\bf{478.95}$ & $82.34\pm0.27$ & $\bf{278.16}$ \\
        $16$ & $69.93\pm0.28$ & $816.07$ & $82.86\pm0.27$ & $499.14$ \\
        $25$ & $\bf{70.47\pm0.28}$ & $1334.27$ & $\bf{82.89\pm0.27}$ & $780.18$ \\
        \bottomrule
    \end{tabular}
\end{table}

\subsection{Visualization of Solved Transport Matrices}\label{subsec:match}

\begin{figure}
    \centering
    \includegraphics[width=\linewidth]{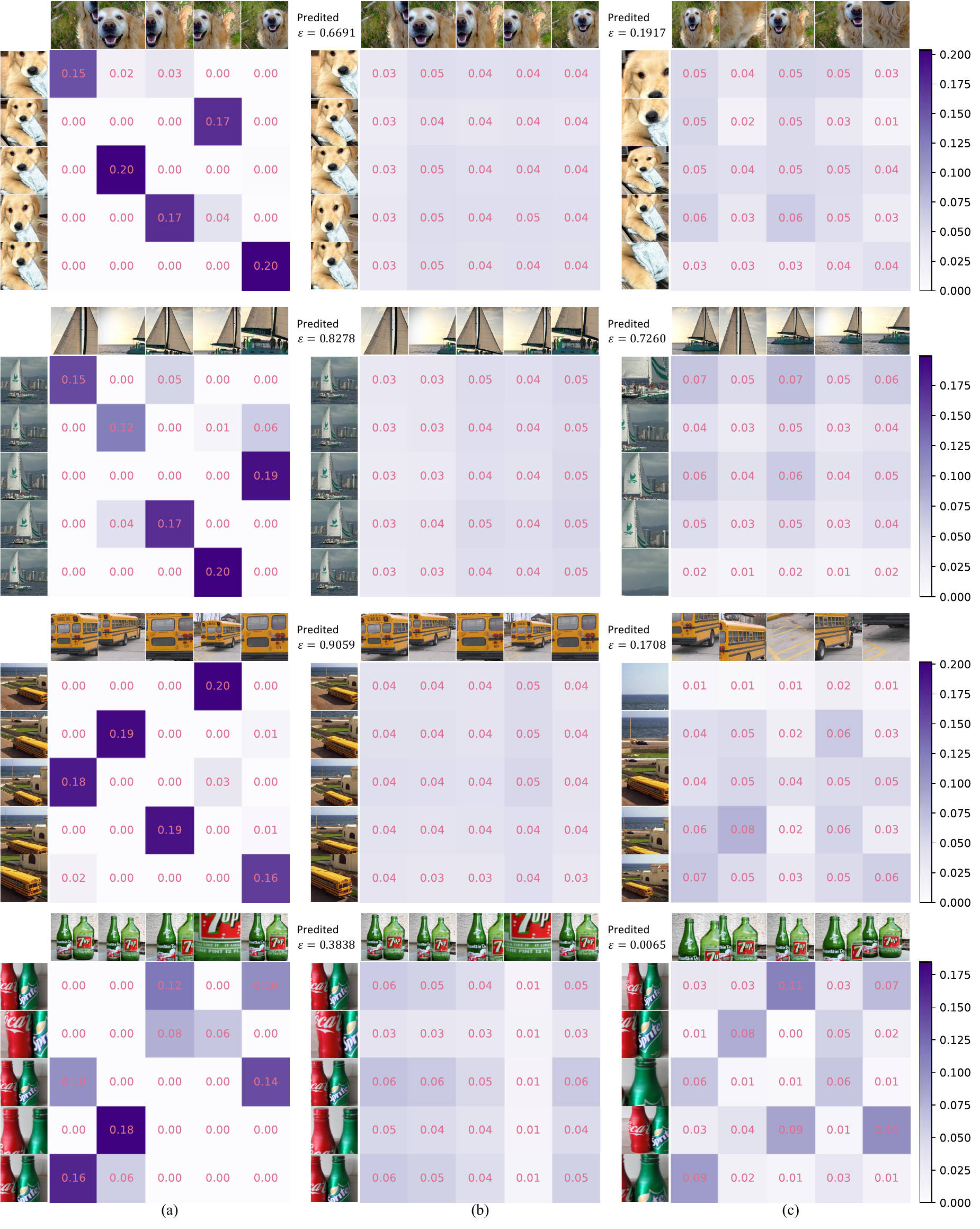}
    \caption{Visualization of solved transport matrices. Results of (a) EMD and (b) Adaptive Metric for sets consisting of similar local features, and the result of (c) Adaptive Metric for sets consisting of dissimilar local features.}
    \label{fig:match_vis_supp}
\end{figure}

For Fig.~\ref{fig:match_vis}, we provide more results in Fig.~\ref{fig:match_vis_supp}.

For sets consisting of similar local patches, the transport matrices solved by EMD (Fig.~\ref{fig:match_vis_supp}~(a))) tend to be very sparse, which is not a desired property because it utilizes equivalent local relations unevenly, neglecting the information provided by lots of important local relations.
In contrast, Adaptive Metric (Fig.~\ref{fig:match_vis_supp}~(b))) generates smoother transport matrices. By allowing ``one-to-many'' matching, it enables a comprehensive utilization of all equivalent local patches and reduces the dependency on a few of them.

By making $\varepsilon$ a learnable parameter, the Modulate Module can control the smoothness of the transport matrix self-adaptively.
For sets consisting of similar local patches, the Modulate Module produces a relatively larger $\varepsilon$, resulting in a smoother transport matrix (Fig.~\ref{fig:match_vis_supp}~(b))).
While for sets consisting of dissimilar local patches, a relatively smaller $\varepsilon$ is predicted, making the transport matrix moderately sparse (Fig.~\ref{fig:match_vis_supp}~(c))).
The Modulate Module makes it possible for our method to handle various set compositions by introducing adaptability into the measure process.

\end{document}